\documentclass[journal,twoside, web]{ieeecolor}
\usepackage{generic}
\usepackage{cite}
\usepackage{amsmath,amssymb,amsfonts}
\usepackage{graphicx}
\usepackage{algorithm,algorithmic,array}
\usepackage{stfloats} % optional, improves *-float placement
\usepackage{tabularx}
\usepackage{tcolorbox}
\usepackage{booktabs}
\usepackage{threeparttable}
\usepackage{multirow}
\usepackage{hyperref}
\usepackage[dvipsnames]{xcolor}
\definecolor{revblue}{RGB}{0,0,180}
\newcommand{\rev}[1]{#1}
\PassOptionsToPackage{dvipsnames}{xcolor}
\usepackage[capitalise]{cleveref}
\hypersetup{hidelinks=true}
\usepackage{textcomp}
\def\BibTeX{{\rm B\kern-.05em{\sc i\kern-.025em b}\kern-.08em
    T\kern-.1667em\lower.7ex\hbox{E}\kern-.125emX}}
\begin{document}

\title{Exploring the Capabilities of Large Language Model Encoders for Image-Text Retrieval in Chest X-rays}

\author{
Hanbin Ko,
Rong Yang,
Gihun Cho,
Inhyeok Baek,
Donguk Kim,
Joonbeom Koo,
Changi Kim,
Dongheon Lee, \IEEEmembership{Member, IEEE},
and Chang Min Park%
\thanks{
This work has been submitted to the IEEE for possible publication.
Copyright may be transferred without notice, after which this version
may no longer be accessible.
}%
\thanks{
Corresponding authors: Dongheon Lee (dhlee13@snu.ac.kr) and
Chang Min Park (morphius@snu.ac.kr).
}%
\thanks{
H. Ko, G. Cho, and C. Kim are with the Interdisciplinary Program in Bioengineering,
Seoul National University Graduate School,
Seoul, Republic of Korea;
and the Integrated Major in Innovative Medical Science,
Seoul National University Graduate School,
Seoul, Republic of Korea
(e-mail: lucasko1994@snu.ac.kr; gihuncho@snu.ac.kr; fr2zyroom@snu.ac.kr).
}%
\thanks{
R. Yang is with the Department of Radiology,
The First Affiliated Hospital, Zhejiang University School of Medicine,
Hangzhou, China
(e-mail: dryangrong@zju.edu.cn).
}%
\thanks{
I. Baek, D. Kim, and J. Koo are with the Seoul National University College of Medicine,
Seoul, Republic of Korea
(e-mail: bih1122@snu.ac.kr; drinkuranium@snu.ac.kr; ciderest@snu.ac.kr).
}%
\thanks{
D. Lee and C. M. Park are with the Department of Radiology,
Seoul National University College of Medicine,
Seoul National University Hospital,
Seoul, Republic of Korea;
the Institute of Medical and Biological Engineering,
Seoul National University Medical Research Center,
Seoul, Republic of Korea;
and the Institute of Radiation Medicine,
Seoul National University Medical Research Center,
Seoul, Republic of Korea
(e-mail: dhlee13@snu.ac.kr; morphius@snu.ac.kr).
}
}

\maketitle

\begin{abstract}
Multimodal learning from paired medical images and clinical text is a central challenge in medical data-driven informatics, where effective cross-modal alignment is critical for scalable analysis and retrieval. In chest radiography, vision--language pretraining is constrained by heterogeneous radiology reports that contain abbreviations, impression-only notes, and institution-specific writing styles. Unlike general-domain settings, naively aggregating large collections of noisy reports can plateau or even degrade multimodal learning when reporting styles differ substantially. We propose a domain-adapted bidirectional large language model text encoder for chest radiograph reports, trained with masked token prediction and supervised contrastive learning on stylistically diverse but clinically equivalent report variants to produce robust, generalizable text embeddings. We then integrate this encoder into a dual-tower contrastive vision--language framework using parameter-efficient adaptation to improve image--text alignment. Across 1.6 million paired studies from public datasets and a de-identified hospital cohort, the proposed models improve bidirectional retrieval accuracy and external generalization, achieving GREEN scores of 0.308 on MIMIC‑CXR and 0.618 on Open‑I, while reducing the degradation observed when abbreviation‑rich, impression‑only hospital reports are added to training.
 \textbf{Significance:} Robust cross-modal embeddings enable scalable retrieval and multimodal representation learning from routine clinical data for biomedical and health informatics.
\end{abstract}


\section{Introduction}
\label{sec:introduction}
Early advances in medical image analysis primarily focused on single-modality tasks such as classification, detection, and segmentation~\cite{shi2021covid,frid2021covid,tabik2020covidgr}. While these models achieved strong performance on curated datasets, they required extensive expert annotations and captured only visual patterns, overlooking the clinical information described in radiology reports. This limitation has motivated a shift toward methods that can jointly learn from visual and textual data.

\emph{Vision-language pretraining (VLP)} has recently emerged as an effective solution for integrating these two modalities. By learning from paired image–text data, VLP models reduce reliance on manual labels and enable tasks such as report generation, retrieval, and cross-modal reasoning. Among these, \textbf{CLIP}~\cite{clip} is particularly influential for its simple contrastive learning framework and strong zero-shot performance, motivating adaptations for radiology~\cite{moon2022multi,chexzero}.

% Adapting CLIP-style frameworks to radiology, however, introduces distinct challenges. Radiology reports differ substantially from natural-language captions, featuring specialized terminology, abbreviations, and linguistic patterns such as negation and uncertainty~\cite{medicalclipsurvey}. \rev{Reports also vary across institutions in section composition (full Findings--Impression pairs versus impression-only summaries), in institution-specific shorthand (e.g., ``BLLF'' for bilateral lower lung field, ``PTX'' for pneumothorax), and in temporal expressions referencing prior studies not visible in the current image~\cite{biovilt,hospital}.} These characteristics hinder the direct transfer of general-domain VLP models and necessitate domain-specific adaptation. As of now, most existing medical VLP systems employ medical specialized BERT-based text encoders~\cite{clinicalbert,bioclinicalbert,biovilt}, and outperform generic encoders. However, they still struggle with the stylistic variability and abbreviation-heavy structure of radiologic reports~\cite{hospital,problem}. \rev{As a consequence, naively aggregating stylistically divergent reports can fail to yield monotonic gains and, as we show in Section~\ref{sec:results}, can even degrade retrieval on cleaner benchmarks when impression-only or abbreviation-heavy corpora are mixed in.}

Adapting CLIP-style frameworks to radiology, however, introduces distinct challenges. Radiology reports differ substantially from natural-language captions, featuring specialized terminology, abbreviations, and linguistic patterns such as negation and uncertainty~\cite{medicalclipsurvey}. \rev{They also vary across institutions in section composition, ranging from full Findings--Impression reports to impression-only summaries, and in institution-specific shorthand, such as ``BLLF'' for bilateral lower lung field and ``PTX'' for pneumothorax. In addition, temporal expressions may refer to prior studies that are not visible in the current image, further complicating image--text alignment~\cite{biovilt,hospital}.} These characteristics hinder the direct transfer of general-domain VLP models and necessitate domain-specific adaptation. Most existing medical VLP systems employ domain-specialized BERT-based text encoders~\cite{clinicalbert,bioclinicalbert,biovilt}, which outperform generic encoders but remain vulnerable to stylistic variability, abbreviation-heavy reporting, and section-level mismatch in radiology reports~\cite{hospital,problem}. \rev{Consequently, simply aggregating stylistically divergent report corpora does not necessarily produce monotonic gains~\cite{cxrclip}; as shown in Section~\ref{sec:results}, adding impression-only or abbreviation-heavy datasets can degrade retrieval performance unless the text encoder is robust to such report-style heterogeneity.}

Large language model (LLM) based embedding frameworks have been reported to provide a stronger foundation for text representation than BERT based encoders~\cite{llm2vec,llm2clip}. These models produce high capacity embeddings that capture nuanced semantic variation across different phrasings~\cite{llm2vec,nvembed}, such as ``no pleural effusion'' and ``pleural spaces are clear,'' while effectively handling longer clinical contexts. These properties make LLM based embeddings well suited for radiologic reports, where clinically equivalent findings are frequently expressed using diverse phrasing and structure. We illustrate this setting using chest X-ray (CXR) imaging, one of the most commonly performed radiologic examinations worldwide, whose reports exhibit substantial variability in structure, length, and expression. \rev{Motivated by these gaps, this study asks whether a domain-adapted bidirectional LLM encoder can provide more robust CXR report representations than BERT-based encoders, and whether such improved text representations can be transferred to a CLIP-style framework to improve clinically faithful image--text retrieval under heterogeneous reporting styles.}

To address this gap, we propose two complementary components: an LLM-based text encoder (\emph{LLM2VEC4CXR}) and a multimodal framework (\emph{LLM2CLIP4CXR}) that integrates the text encoder with a vision encoder for improved image–text alignment.

\textbf{(1) LLM2VEC4CXR:} a domain-adapted LLM encoder for CXR reports that is robust to abbreviation-heavy and stylistically heterogeneous clinical language. It combines masked token prediction, supervised contrastive learning, and latent attention pooling to produce stable, clinically meaningful text embeddings across both Findings and Impression sections.

\textbf{(2) LLM2CLIP4CXR:} a multimodal framework that integrates LLM2VEC4CXR with a vision encoder for image–text alignment. Using LoRA-based parameter-efficient fine-tuning, it transfers improved text understanding to the image domain, enhancing retrieval accuracy and cross-dataset generalization.

Trained on \textbf{1.6M CXRs} from public and private sources, our models outperform BERT-based and prior medical CLIP frameworks on both standard and clinically oriented metrics. Together, these results emphasize that advancing clinical text comprehension is more decisive for multimodal generalization than simply increasing data volume.

\begin{figure*}[h]
\centering
\includegraphics[width=\linewidth]{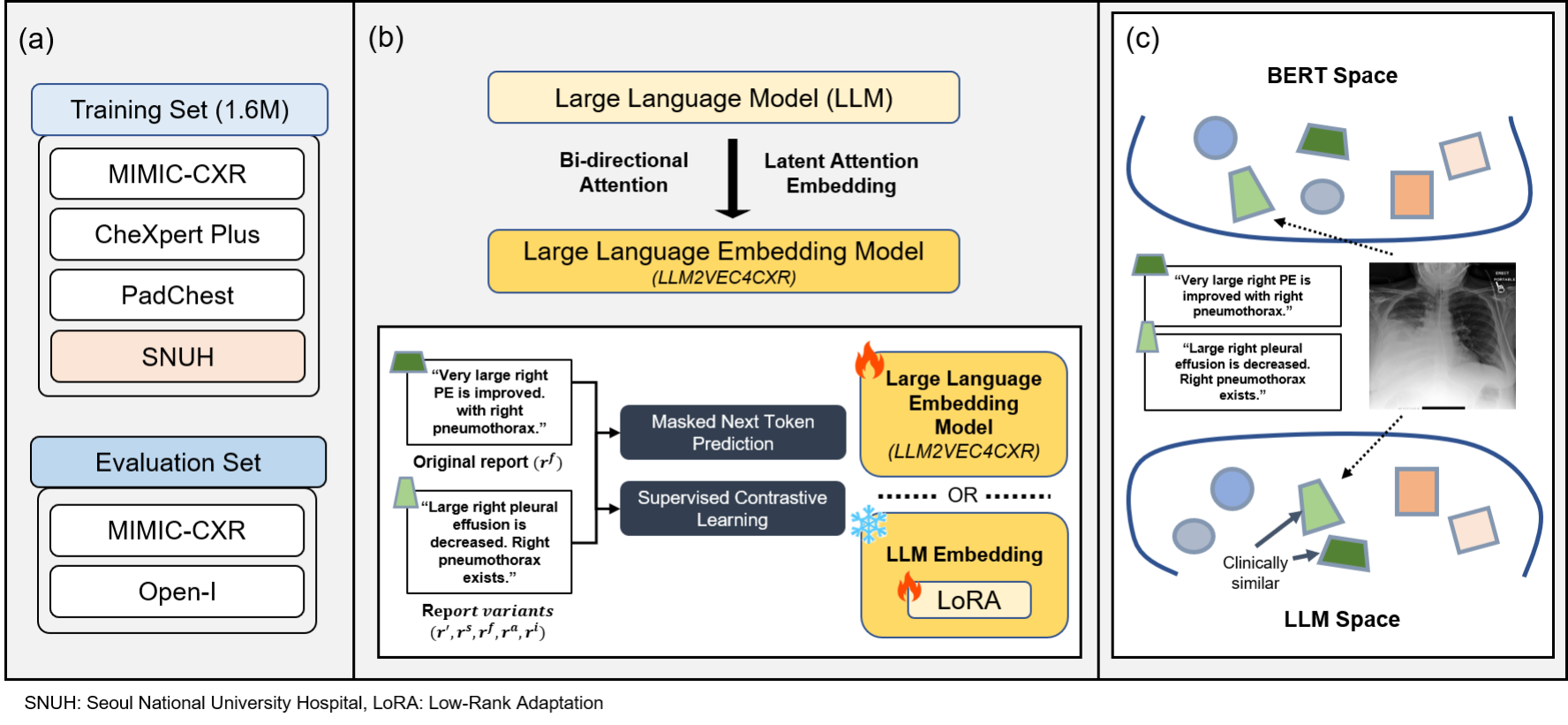}
\caption{
Overview of the proposed framework. 
(a) Training and evaluation datasets combining public and private data for 1.6M CXR image–report pairs. 
(b) \textbf{LLM2VEC4CXR}: domain-adapted LLM trained with masked token prediction and contrastive learning. 
(c) Compared to BERT, LLM-based embeddings cluster clinically similar phrases, yielding more robust representations for multimodal learning.
}

\label{fig:overview}
\end{figure*}

\section{Related works}

Our work builds on two primary research areas: medical vision-language
modeling and the use of large language models as text encoders. We review both
areas and then position our contributions within this context.

\subsection{Medical vision-language pretraining}
Contrastive learning on paired images and text, as introduced by CLIP~\cite{clip}, has proven effective for transferable multimodal representations. Several medical adaptations follow this paradigm, including ConVIRT~\cite{convirt}, CheXzero~\cite{chexzero}, BioViL/BioViL-T~\cite{biovil,biovilt}, and MedCLIP~\cite{wang2022medclip}, which pair chest X-rays with reports in CLIP-style frameworks. These models typically replace CLIP's text encoder with biomedical BERT variants such as BioClinicalBERT~\cite{bioclinicalbert}, PubMedBERT~\cite{zhang2023biomedclip}, or CXR-BERT~\cite{biovilt}. While these encoders improve biomedical language coverage, they can still be challenged by sectioned report structure, institution-specific abbreviations, negation, uncertainty, and the semantic relationship between \emph{Findings} and \emph{Impression} sections. Thus, the text encoder remains a key bottleneck in medical VLP.

% \rev{A related line of work focuses on CXR report generation, where vision backbones are coupled with language decoders to synthesize radiology reports, as in MAIRA-2~\cite{maira2}, CheXagent~\cite{chen2024chexagent}, CXR-LLAVA~\cite{cxrllava}, and the abnormality-guided META-CXR~\cite{edirisinghe2025chest}. These studies primarily use language models on the output side for report generation. In contrast, our work adapts an LLM on the input side as a domain-specialized bidirectional text encoder for image--text retrieval and multimodal alignment.}

\rev{CXR report generation represents a related but distinct line of work~\cite{nimalsiri2023automated}. Conventional transformer-based and recent LLM/VLM-based systems typically couple visual encoders with autoregressive language decoders to synthesize free-text radiology reports, as in MAIRA-2~\cite{maira2}, CheXagent~\cite{chen2024chexagent}, CXR-LLAVA~\cite{cxrllava}, and META-CXR~\cite{edirisinghe2025chest}. These methods primarily optimize image-to-report decoding, whereas our study addresses the input-side representation problem by adapting an LLM into a bidirectional, domain-specialized report encoder for image--text retrieval and multimodal alignment under heterogeneous CXR reporting styles.}

\subsection{Evaluation challenges in CXR retrieval}
Most clinical CLIP retrieval studies report recall at top-$k$~\cite{convirt,cxrclip},
which measures exact string matching between queries and reports. Yet radiology
reports frequently contain multiple semantically equivalent phrasings, so
retrieving a clinically correct but textually different report is penalized as
an error. To address this, radiology report generation research
~\cite{zhang2024rexrank,maira2,chen2024chexagent} has proposed clinically
oriented metrics such as CheXbert F1~\cite{chexbert}, RadGraph
F1~\cite{radgraph}, RaTEScore~\cite{ratescore}, and GREEN~\cite{ostmeier2024green}, which assess whether
retrieved text captures the correct entities and relations. We adopt this
direction, applying clinically relevant metrics to retrieval and focusing on
\emph{clinical faithfulness} rather than surface-level similarity.

\subsection{LLMs as text encoders}
Beyond generation, LLMs are increasingly adapted for encoding tasks. Methods such as LLM2VEC~\cite{llm2vec} and NV-Embed~\cite{nvembed} show that LLM representations can outperform specialized encoders when fine-tuned for embeddings. Their use has expanded to domains such as table analysis~\cite{koloski2025llm} and recommendation systems~\cite{zhang2024notellm}. Early biomedical efforts include BMRetriever~\cite{bmretriever}, which tunes LLMs for clinical retrieval, but applications to multimodal medical tasks remain limited. In radiology, there has been little exploration of LLM encoders for chest X-ray report alignment, despite their potential to capture semantic variation across diverse reporting styles.

\subsection{Positioning and contributions}
In summary, prior studies have primarily adapted BERT-style text encoders for medical CLIP frameworks, but these models still lack strong clinical alignment and rely heavily on recall-based evaluation. Moreover, most existing work has been trained on curated public corpora, whereas real-world private reports are substantially noisier, abbreviated, and often restricted to shorter sections such as \emph{Impression} notes. 

Our work addresses these limitations through \emph{LLM2VEC4CXR} and  \emph{LLM2CLIP4CXR}, which leverage LLM encoders to capture richer clinical semantics and enable more robust image-text alignment. Together, they advance medical vision-language pretraining by improving cross-dataset generalization and enhancing the clinical relevance of retrieval performance.

\section{Methods}
\label{sec:method}

We introduce two main models: \emph{LLM2VEC4CXR}, a specialized LLM encoder
for CXR reports, and \emph{LLM2CLIP4CXR}, which integrates this encoder
with a vision backbone in a two-tower CLIP-style framework. An overview of the proposed framework is shown in \cref{fig:overview}.

\subsection{LLM2VEC4CXR}

CXR reports contain long sentences, abbreviations, temporal references, and
variable formatting, which make standard BERT encoders brittle.   \emph{LLM2VEC4CXR} addresses the complexity of radiology language. Following
LLM2VEC~\cite{llm2vec}, we remove the causal attention mask from
decoder-only architectures to enhance bidirectional context encoding, and adopt
latent attention pooling (LAP) for more informative
global embeddings. The model is trained with two objectives:
Masked token prediction (MTP) and supervised contrastive learning.

\subsubsection{Data generation}
To expose the model to clinically equivalent but stylistically diverse inputs,
we generate multiple variants of each report ($r$) using 
\emph{Gemini2.0-Flash}~\cite{gemini} and 
\emph{Deepseek-R1-Distill-Qwen-14B}~\cite{deepseekr1}. 
These include:
\begin{itemize}
    \item \textbf{Clinically similar reports ($r'$):} Rephrasings that preserve meaning while altering style.
    \item \textbf{Sentence splitting ($r^s$):} Decomposition of multi-finding sentences into atomic observations~\cite{maira2}.
    \item \textbf{Omitting temporal references ($r^o$):} Removal of temporal expressions and change descriptors.
    \item \textbf{Anatomical partitioning ($r^d$):} Segmentation by anatomical region, followed by structured recombination.
    \item \textbf{Summarization pairs:} Linking \emph{Findings} ($r^f$) with \emph{Impression} ($r^i$) sections.
\end{itemize}

\rev{To assess whether these synthetic variants were clinically valid as positive pairs, we performed an expert review of 50 source--synthetic report pairs sampled from the training-positive pool. A thoracic radiologist with approximately 15 years of clinical experience assessed whether each synthetic report preserved the current clinically relevant content, including findings, negation or uncertainty, anatomical location, laterality, and severity. Each pair was categorized as clinically acceptable or clinically unacceptable. The radiologist judged 46 of 50 pairs (92\%) as clinically acceptable positive pairs for contrastive training, supporting the use of these synthetic variants as training positives.}

All variants also participate in MTP pretraining, strengthening robustness to
heterogeneous report styles. To capture clinical shorthand, we additionally
include MIMIC's \emph{Indication} fields, which contain frequent acronyms. 
A summary is given in \Cref{tab:mntp}.

% \begin{figure*}[h]
% \centering
% \includegraphics[width=\linewidth]{figures/report_sample.pdf}
% \caption{Examples of report variations used for training. Variants include
% rephrasings ($r'$), sentence splits ($r^s$), omission of temporal references
% ($r^o$), anatomical partitioning ($r^d$), and summarization pairs
% (\emph{Findings} $r^f$ to \emph{Impression} $r^i$).}
% \label{fig:ex}
% \end{figure*}

\subsubsection{Bidirectional adaptation of a decoder-only LLM}
Let a report be tokenized into a sequence $\mathbf{x}=(x_1,\dots,x_L)$.
We convert a decoder-only transformer into a bidirectional encoder by removing
the causal (triangular) attention mask. Concretely, in each self-attention layer
we compute
\begin{equation}
\mathrm{Attn}(\mathbf{Q},\mathbf{K},\mathbf{V})
= \mathrm{softmax}\!\left(\frac{\mathbf{Q}\mathbf{K}^\top}{\sqrt{d}} + \mathbf{M}\right)\mathbf{V},
\end{equation}
where $d$ is the head dimension and $\mathbf{M}$ is an attention mask.
In a standard decoder, $\mathbf{M}=\mathbf{M}^{\mathrm{causal}}$ with
$(\mathbf{M}^{\mathrm{causal}})_{ij}=0$ if $j\le i$ and $-\infty$ otherwise.
We set $\mathbf{M}=\mathbf{0}$ to allow each token to attend to both preceding
and following context, producing bidirectional hidden states
$\mathbf{H}=f_{\theta}(\tilde{\mathbf{x}})\in\mathbb{R}^{L\times d_{\text{model}}}$.
\rev{This conversion is required because the decoder-only LLM is used as a report encoder rather than as an autoregressive generator. For CXR reports, full-context attention is appropriate because findings are commonly organized by anatomy rather than by a fixed causal order, so the final report embedding should integrate observations, modifiers, and section-level information across the entire report.}

\subsubsection{Masked token prediction}
We pretrain the bidirectional text encoder using masked token prediction (MTP), following LLM2VEC~\cite{llm2vec}. Given a report, a subset of tokens is replaced with a mask symbol, and the model is trained to recover the original tokens from bidirectional context:
\begin{equation}
\mathcal{L}_{\mathrm{MTP}}
= -\frac{1}{|\mathcal{M}|}\sum_{i\in\mathcal{M}}
\log p_\theta(x_i \mid \tilde{\mathbf{x}}).
\end{equation}
Because the causal mask is removed, masked tokens are predicted from both left and right context, encouraging global report understanding rather than next-token continuation.

\subsubsection{LAP for global text embeddings}
For contrastive learning, token-level hidden states must be converted into a single report embedding. We use latent attention pooling (LAP)~\cite{nvembed}, in which learnable latent queries attend to token representations and produce a pooled report vector. The pooled vector is passed through a projection head and $\ell_2$-normalized before contrastive learning. Compared with mean pooling, LAP allows the model to learn which report tokens contribute most to the global clinical representation.

\subsubsection{Supervised contrastive learning on report variants}
After MTP pretraining, we fine-tune the encoder with supervised contrastive learning
to cluster clinically equivalent texts (e.g., original reports and their generated variants)
while separating unrelated reports.
Let a mini-batch contain $B$ input texts, each mapped to a normalized embedding
$\{\mathbf{u}_i\}_{i=1}^B$ by the text encoder and projection head. Each sample $i$ has a supervision label $y_i$ indicating membership in a positive group
(e.g., same underlying report instance, same finding label, or an instruction-aligned pair).
Define the positive set for anchor $i$ as
\begin{equation}
\mathcal{P}(i) = \{\,p\in\{1,\dots,B\}\setminus\{i\}\;|\;y_p=y_i\,\}.
\end{equation}
Using cosine similarity (equivalently dot product of unit vectors),
$\mathrm{sim}(\mathbf{u}_i,\mathbf{u}_j)=\mathbf{u}_i^\top\mathbf{u}_j$,
the supervised contrastive (InfoNCE-style) loss is
\begin{align}
\mathcal{L}_{\mathrm{C}}
&=
-\frac{1}{B}\sum_{i=1}^{B}
\frac{1}{|\mathcal{P}(i)|}
\sum_{p\in\mathcal{P}(i)}
\log
\frac{
\exp\!\big(\mathbf{u}_i^\top\mathbf{u}_p/\tau\big)
}{
\displaystyle
\sum_{a\in\{1,\dots,B\}\setminus\{i\}}
\exp\!\big(\mathbf{u}_i^\top\mathbf{u}_a/\tau\big)
}
\end{align}

where $\tau>0$ is a temperature hyperparameter.
This objective explicitly pulls together embeddings of clinically consistent
rephrasings/sections while pushing apart embeddings from different studies.

\subsection{LLM2CLIP4CXR} 
\label{subsec:llm2clip4cxr}

\emph{LLM2CLIP4CXR} integrates the domain-adapted text encoder 
\emph{LLM2VEC4CXR} with a vision encoder in a dual-tower CLIP-style framework. 
The vision tower encodes the CXR image and the text tower encodes the paired report, 
with both embeddings projected into a shared normalized representation space.

Given a batch of image--report pairs, we optimize the standard symmetric CLIP contrastive loss, which aligns matched image--report pairs and separates mismatched pairs within the batch. During training, the vision encoder is fully optimized, the projection layers of both encoders are trained from scratch, and the text encoder is adapted using LoRA-based updates. This design transfers clinically informed report representations into the multimodal space while keeping adaptation of the LLM backbone efficient.

\section{Experiments}
\label{sec:experiments}

We describe the datasets, implementation details, preprocessing steps, and evaluation methodology used in our experiments.

\subsection{Datasets}

\emph{LLM2VEC4CXR} is trained on the training splits of MIMIC~\cite{mimic3}, CheXpert-plus~\cite{chexpertplus}, and the generated variants described in \Cref{tab:mntp}. For \emph{LLM2CLIP4CXR}, we start with paired samples from MIMIC-CXR and incrementally add CheXpert-plus, PadChest~\cite{padchest}, and a dataset from Seoul National University Hospital (SNUH; IRB no. 2405-070-1536), resulting in a combined training set of around 1.6M studies. Only frontal X-rays are used. To exploit the LLM’s ability to handle longer text sequences, we retain full \emph{Findings} or \emph{Impression} sections when constructing non-sectioned training corpora. Notably, these datasets differ substantially in reporting style: MIMIC provides full structured reports, PadChest and SNUH contain impression-only summaries, and SNUH reports often consist of abbreviated notes. This heterogeneity provides a natural testbed for robustness to noise and style variation.

\begin{table}[h]
\centering
\caption{Counts of preprocessed reports used for LLM2VEC4CXR training.}
\label{tab:mntp}
\begin{tabular}{lc}
\toprule
\textbf{Pair Type} & \textbf{Count} \\
\midrule
Original reports & 319,564 \\
Split reports & 178,993 \\
Prior-omitted reports & 169,491 \\
Anatomical-partitioned reports & 178,096 \\
Clinically similar reports & 272,230 \\
\midrule
\textbf{Total} & 1,118,374 \\
\bottomrule
\end{tabular}
\end{table}

We evaluate on two datasets: a held-out MIMIC-CXR internal validation split that is disjoint from training and tuning at the study level, and the \emph{Open-I}~\cite{openi} external validation set. To reduce
temporal ambiguity, we exclude Open-I reports containing temporal expressions.
The exclusion keywords are adopted from BioViL-T~\cite{biovilt}, but we apply
them only as a filter: any report containing one or more of these keywords is
removed from the test set. We then retain a single comprehensive report per
study to avoid multiple time points. Dataset statistics are summarized in
\Cref{tab:dataset_stats}.

\begin{table}[h]
\centering
\caption{Overview of the datasets.}
\label{tab:dataset_stats}
\begin{tabular}{lcccc}
\toprule
\textbf{Dataset} & \textbf{Train} & \textbf{Tune} & \textbf{Int.\ Val.} & \textbf{Ext.\ Val.} \\
\midrule
MIMIC-CXR~\cite{mimic3}     & 247{,}648 & 2{,}032 & 3{,}394 & -- \\
CheXpert-plus~\cite{chexpertplus} & 190{,}668 & 202     & --      & -- \\
PadChest~\cite{padchest}      & 60{,}261  & 1{,}604 & --      & -- \\
Open-I~\cite{openi}        & --        & --      & --      & 1{,}825 \\
SNUH          & 1{,}136{,}410 & 1{,}328 & --   & -- \\
\bottomrule
\end{tabular}
\vspace{2pt}
{\footnotesize
\emph{Abbreviations:} Int.\ Val., internal validation; Ext.\ Val., external validation;  SNUH, Seoul National University Hospital.
}
\end{table}

\subsection{Additional input processing}
\label{sec:sectioned_processing}
To improve section awareness, we use two complementary input-side mechanisms. First, we prepend inline section placeholders (\texttt{[FINDINGS]}, \texttt{[IMPRESSION]}) to the corresponding text span when training section-aware variants of \emph{LLM2CLIP4CXR}. \rev{Second, we prepend a fixed task-type instruction prompt indicating the training objective (semantic similarity, summarization, or finding-level classification; see Table~\ref{tab:prompt_templates}). The two mechanisms are orthogonal -- placeholders mark report sections, prompts mark the training objective --- and may co-occur in a single input. This section-aware prompting was particularly important for integrating datasets with impression-only reports (e.g., PadChest, private corpus), where otherwise the model could underfit detailed findings and overfit condensed styles.}

\subsection{Implementation details}

\subsubsection{Model architecture}
For the text tower, we build on the LLM2VEC~\cite{llm2vec} encoder and
construct \emph{LLM2VEC4CXR} in two variants: (i) a
\emph{LoRA-based parameter-efficient update}, where only low-rank adapters are
trained while keeping the backbone frozen, and (ii) a \emph{fully fine-tuned
version}, where all model parameters are updated. For
\emph{LLM2CLIP4CXR}, we extend the text encoder with an additional LoRA
layer and a projection head. The vision tower is an EVA02-CLIP-L/14 backbone,
resized for $448 \times 448$ pixels, with its own projection head of dimension
$1280$. Both projection layers are trained from scratch.

As a baseline, we trained a standard BERT~\cite{bert} model under the same settings and integrated it into a CLIP-style framework, referred to as \emph{BERT4CXR} and \emph{CLIP4CXR}, respectively. Our \emph{LLM2VEC4CXR} models use either
\emph{Llama3.2-1B} or \emph{Llama3-8B}~\cite{dubey2024llama} as backbone LLMs.
All standard medical CLIP variants are retrained on our dataset for consistent
comparison.

\subsubsection{Training configuration}
All models are trained on four NVIDIA A6000 GPUs. \emph{LLM2VEC4CXR} is
pretrained with MTP for one epoch and further optimized with supervised
contrastive learning. Latent attention pooling~\cite{nvembed} is used to form
global text embeddings. CLIP training configurations follow
LLM2CLIP~\cite{llm2clip} with a per-GPU batch size of 256. \rev{A computational cost comparison of the 1B and 8B LLM2VEC4CXR variants is provided in Appendix~\ref{app:compute_tradeoff}.} All experiments were
implemented in PyTorch~(v2.4.1) using Python~3.8. More details on model
configurations and versions are provided in our public
repository: \url{https://github.com/lukeingawesome/llm2clip4cxr.git}.

\subsection{Text-only evaluation}
\label{sec:text_evaluation}

To evaluate the text encoder independently from the vision backbone, we designed five text-only tasks that probe different aspects of linguistic and clinical robustness. These tasks collectively assess the model’s ability to handle reporting style variation, detect subtle semantic inconsistencies, and recognize clinically equivalent expressions across datasets. Each task isolates a specific challenge commonly encountered in radiology reporting—such as abbreviation comprehension, omission of temporal context, or paraphrased impressions—to provide a fine-grained analysis of text understanding. A detailed overview of the five evaluation tasks is provided in \Cref{tab:text_tasks}.

\subsection{Multimodal evaluation}
For \emph{LLM2CLIP4CXR}, we align frontal CXRs with reports. We first evaluate
Top-$k$ retrieval within each test set. To assess clinical correctness, we
further apply the same metrics used in Task~5 (CheXbert F1, RadGraph F1,
RaTEScore, and GREEN) when retrieving reports from the MIMIC
train/validation pool. This avoids biases introduced by external datasets whose
test reports may not cover all clinical variations.

Automated metrics may miss clinically important aspects of report quality.
To complement them, we conducted a qualitative ranking study on 200 Open-I cases with three medical students (3, 8, and 44 months of training) and four large LLMs
(\emph{GPT-4o}~\cite{hurst2024gpt}, \emph{Gemini 1.5 Pro}~\cite{gemini}, \emph{DeepSeek-V3}~\cite{deepseek}, and
\emph{DeepSeek-R1}~\cite{deepseekr1}). We excluded half of the normal studies to ensure case diversity and compared ground-truth reports with retrieved and generated outputs. Candidate outputs were randomized to avoid order bias. Raters were instructed to rank the candidate reports by clinical accuracy relative to the ground truth, with ties permitted. All raters followed the LLM-RadJudge~\cite{judge} protocol, which prioritizes critical clinical findings over minor or irrelevant details.

\rev{To anchor this qualitative evaluation to expert clinical judgment, a thoracic radiologist additionally reviewed a 72-case subset sampled from the same 200 Open-I cases. To reduce expert-rater burden while preserving representative model coverage, this subset included three systems: \emph{LLM2CLIP4CXRall}, representing LLM-encoder retrieval; BC2C~\cite{bc2c}, representing BERT-based CLIP retrieval; and MAIRA-2~\cite{maira2}, representing report generation. The radiologist used the same randomized ranking protocol.}

\begin{table*}[h]
\centering
\caption{Overview of text-only evaluation tasks for assessing LLM2VEC4CXR. Top-$k$ retrieval is used unless otherwise noted.}
\label{tab:text_tasks}
\resizebox{\textwidth}{!}{
\begin{tabular}{p{1.0cm}p{3cm}p{6.5cm}p{4.5cm}}
\toprule
\textbf{Task} & \textbf{Name} & \textbf{Description} & \textbf{Metric(s)} \\
\midrule
1 & Prior-omitted - Original matching & Given a prior-omitted report ($r^o$), the model retrieves the original report to evaluate sensitivity to omitted context. & Top-$k$ retrieval \\
\addlinespace[2pt]
2 & Report summarization & Given the \emph{Findings} section, the model retrieves the corresponding \emph{Impression}, assessing summarization and abstraction ability. & Top-$k$ retrieval \\
\addlinespace[2pt]
3 & Report error discrimination & Following \emph{ReXErr}~\cite{rao2024rexerr}, three erroneous impressions (false negation, severity change, etc.) are synthesized per true report. The model must identify the correct impression given the \emph{Findings} section. & Accuracy \\
\addlinespace[2pt]
4 & Medical acronym understanding & Reports containing acronyms (e.g., ``BLLF'', ``PTX'') are paired with expert-refined, fully expanded versions (e.g., ``bilateral lower lung field''). Tests robustness to lexical variation. & Top-$k$ retrieval \\
\addlinespace[2pt]
5 & Clinical similarity matching & Given a findings section from Open-I, the model retrieves the most similar MIMIC-CXR findings report to assess clinical equivalence. & RadGraph F1, CheXbert F1, RaTEScore, GREEN \\
\bottomrule
\end{tabular}
}
\end{table*}

\section{Results}
\label{sec:results}
% Main quantitative and qualitative results. Refer to figures and tables.

We report results for both the \textbf{text-only} setting, implemented using  \emph{LLM2VEC4CXR}, and the \textbf{multimodal} setting, implemented using  \emph{LLM2CLIP4CXR}. The evaluation focuses on retrieval performance and clinically oriented measures. Overall, the LLM-based text encoders demonstrate stronger semantic alignment and generalizability than BERT-based and existing CLIP-style approaches, with consistent improvements on external validation.

\subsection{Text-only results}
\Cref{tab:llm2vec_results} compares \emph{LLM2VEC4CXR} with baseline text
encoders across the five evaluation tasks. Baselines include generic
BERT~\cite{bert}, medical BERT variants
(\emph{PubMedBERT}~\cite{zhang2023biomedclip},
\emph{BioClinicalBERT}~\cite{bioclinicalbert},
\emph{CXR-BERT}~\cite{biovilt}, \emph{ClinicalBERT}~\cite{clinicalbert}), and
two general-domain LLM encoders (\emph{LLM2VEC} 1B and 8B). We also evaluate our proposed variants, \emph{LLM2VEC4CXR} (1B/8B) and \emph{LLM2VEC4CXR+} (fully fine-tuned), under the same protocol.

\begin{table*}[h]
\centering
\caption{Text-only evaluation of \emph{LLM2VEC4CXR} and baselines. 
Tasks 1–2: Top-$k$ retrieval, Task~3: accuracy, Task~4: Top-$k$ retrieval, 
Task~5: clinical efficacy metrics.}
\label{tab:llm2vec_results}
\begin{threeparttable}
\resizebox{\textwidth}{!}{
\begin{tabular}{l|ccc|ccc|c|ccc|cccc}
\toprule
 & \multicolumn{3}{c|}{\textbf{\shortstack{Prior-omitted \\ Original matching}}} & \multicolumn{3}{c|}{\textbf{\shortstack{Report \\summarization}}} & \textbf{\shortstack{Report error \\ discrimination}} &\multicolumn{3}{c|}{\textbf{\shortstack{Medical acronym \\ understanding}}}& \multicolumn{4}{c}{\textbf{\shortstack{Clinical similarity \\matching}}}\\
\textbf{Model} & @1 & @5 & @10 & @1 & @5 & @10 & Acc & @1 &@3 & @5 & RadGraph & MF1 & RaTE & GREEN\\
\midrule
\textbf{Base-BERT}~\cite{bert}       & 0.520 & 0.664 & 0.708 & 0.016 & 0.039 & 0.054 & 0.092 & 0.137 & 0.384 & 0.513 & 0.324 & 0.206  & 0.693 & 0.632\\
\textbf{PubMedBERT}~\cite{zhang2023biomedclip}      & 0.447 & 0.585 & 0.638 & 0.030 & 0.066 & 0.097 & 0.223 & 0.274& 0.470 & 0.598 & 0.309 & 0.227  & 0.696 & 0.638 \\
\textbf{BioClinicalBERT}~\cite{bioclinicalbert}  & 0.543 & 0.690 & 0.743 & 0.017 & 0.038 & 0.054 & 0.100& 0.231 & 0.385 & 0.556 & 0.298 & 0.207  & 0.681 & 0.616\\
\textbf{ClinicalBERT}~\cite{clinicalbert}  & 0.546 & 0.711 & 0.767 & 0.016 & 0.039 & 0.054 & 0.110 &0.376 & 0.692 & 0.795 & 0.336 & 0.284   & 0.711 & 0.665\\
\textbf{CXR-BERT-Specialized}~\cite{biovilt}      & 0.395 & 0.537 & 0.600 & 0.082 & 0.155 & 0.196 & 0.484& 0.248 & 0.513 & 0.650 & 0.284 & 0.297  & 0.678 & 0.664\\
\hline
\textbf{BERT4CXR} & 0.786 & 0.821 & 0.875 & 0.048 & 0.093 & 0.118 & 0.419 & 0.385 &0.710&0.804& 0.313& 0.275 &  0.708& 0.642\\ 
\textbf{LLM2VEC(1B)}~\cite{llm2clip} & 0.576 & 0.766 & 0.826 & 0.053 & 0.093 & 0.116 & 0.552 & 0.368 & 0.675 & 0.769 & 0.392 & 0.682  & \underline{0.724}& 0.662 \\ 
\textbf{LLM2VEC(8B)}~\cite{llm2clip} & 0.703 & 0.823 & 0.871 & 0.090 & 0.145 & 0.172 & 0.637 &0.368&0.735&0.846&\textbf{0.404}& 0.689  & 0.722& 0.667\\ 
\textbf{LLM2VEC4CXR(1B)} & 0.894 & 0.912 & 0.918 & 0.162 & 0.225 & 0.261 & 0.672 &0.556&0.795&0.870& 0.396 & 0.705  & 0.723 & 0.653 \\
\textbf{LLM2VEC4CXR(8B)} & \textbf{0.933} & \textbf{0.989} & \textbf{0.994} & \textbf{0.212} & \textbf{0.305} & \textbf{0.352} & \textbf{0.841}&\textbf{0.611}&\textbf{0.875}&\textbf{0.926}&\underline{0.402}& \underline{0.712}  & \textbf{0.727} & \underline{0.676} \\
\textbf{LLM2VEC4CXR+(1B)} & \underline{0.918} & \underline{0.984} & \underline{0.989} & \underline{0.203} & \underline{0.295} & \underline{0.339} & \underline{0.826}&\underline{0.607}&\underline{0.854}&\underline{0.910}&0.373& \textbf{0.715} & \underline{0.724} & \textbf{0.686} \\
\bottomrule
\end{tabular}
}
\begin{tablenotes}
\footnotesize
\item \textbf{Bold} indicates the best result; \underline{underline} indicates the second-best result.
\end{tablenotes}
\end{threeparttable}
\end{table*}

\subsubsection{Task 1: Prior-omitted Original matching}
\emph{LLM2VEC4CXR} variants reach near-perfect top-$5$/top-$10$ performance.
Top-1 exceeds $0.9$ for the 8B and fully fine-tuned models, clearly surpassing
\emph{BERT4CXR} under the same protocol.

\subsubsection{Task 2: Report summarization}
General \emph{LLM2VEC} already surpasses all BERT baselines, which struggle to
link \emph{Findings} and \emph{Impression}, especially under style shifts in
Open-I. Domain adaptation in \emph{LLM2VEC4CXR} yields further gains, showing
that LLM encoders capture summary relationships more reliably.

\subsubsection{Task 3: Error discrimination}
Most BERT variants operate below chance ($\approx 0.25$) while CXR-BERT performs slightly better. In contrast,
\emph{LLM2VEC4CXR+} attains $0.826$ accuracy and  \emph{LLM2VEC4CXR (8B)} reaches $0.841$.
Notably, even the \emph{LLM2VEC (1B)} model, without radiology-specific tuning, surpasses all BERT
baselines. These results indicate that LLM-derived embeddings more reliably encode contradictions
and subtle semantic edits (e.g., false negation, severity/location changes), enabling robust detection
of clinically erroneous or different text.

\subsubsection{Task 4: Acronym Comprehension} BERT-based encoders frequently fail to expand acronyms, whereas
\emph{LLM2VEC} and \emph{LLM2VEC4CXR} both excel. Notably, the general
\emph{LLM2VEC} performs on par with BERT4CXR despite no explicit exposure to
medical abbreviations.

\subsubsection{Task 5: Clinical similarity}
In cross-dataset matching, LLM-based encoders consistently outperform BERT
variants on clinically oriented metrics. Domain-adapted \emph{LLM2VEC4CXR}
further improves alignment. A small drop in RadGraph F1 with full fine-tuning
suggests sensitivity to local structure; however, GREEN remains high, indicating
preservation of overall clinical correctness. Qualitative examples in
\Cref{tab:retrieval_example_openi} illustrate that \emph{LLM2VEC4CXR}
retrieves key findings reliably, whereas BERT-based models often hallucinate or
misstate details.

\begin{table*}[h]
\centering
\footnotesize
\caption{Task~5 qualitative example: Open-I query vs.\ top-1 retrieved MIMIC report.}
\label{tab:retrieval_example_openi}
\setlength{\tabcolsep}{6pt}
\renewcommand{\arraystretch}{1.15}
\begin{threeparttable}
\begin{tabularx}{\textwidth}{
>{\raggedright\arraybackslash}p{0.18\textwidth}
>{\raggedright\arraybackslash}X
}
\toprule
\textbf{Model} & \textbf{Text} \\
\midrule
Ground Truth (Query) &
\textcolor{blue}{Cardiomegaly} is noted, with central pulmonary vascular prominence and coarsened interstitial markings,
\textcolor{blue}{suspicious for developing interstitial pulmonary edema}.
No focal consolidation, pneumothorax, or definite effusion identified.
No acute bony abnormality seen. \\
\addlinespace
Base BERT &
\textcolor{red}{Patient is status post median sternotomy and CABG.}
Heart size is normal.
Mediastinal contour is unremarkable.
\textcolor{red}{Fullness of the right hilum likely reflects known lung cancer.}
\textcolor{red}{Mild upper zone}
\textcolor{blue}{pulmonary vascular redistribution suggests mild pulmonary vascular congestion.}
\textcolor{red}{Coarse interstitial opacities are noted in the lung bases which may reflect chronic interstitial abnormality.
Additionally, more focal opacity is seen within the periphery of the right lung base, a nonspecific finding.}
No pleural effusion or pneumothorax is identified.
No acute osseous abnormalities seen. \\
\addlinespace
ClinicalBERT &
\textcolor{red}{Heart size is normal.
Mediastinal and hilar contours are unchanged with architectural distortion, superior bilateral hilar retraction,
traction bronchiectasis, nodularity and fibrosis involving both upper lobes.}
No new focal consolidation, pleural effusion or pneumothorax is present.
\textcolor{orange}{No pulmonary edema is detected, though there may be mild pulmonary vascular congestion.}
No acute osseous abnormality is visualized. \\
\addlinespace
LLM2VEC (1B) &
The heart is \textcolor{orange}{severely} \textcolor{blue}{enlarged}.
There are \textcolor{orange}{perihilar hazy opacities} with vascular indistinctness
\textcolor{blue}{compatible with mild to moderate pulmonary edema}.
\textcolor{red}{Assessment of the lung bases is limited due to technique, but there may be atelectasis.}
No large pleural effusion or pneumothorax is seen. \\
\addlinespace
LLM2VEC4CXR+ &
The heart size is \textcolor{orange}{borderline} \textcolor{blue}{enlarged}.
Mediastinal contours are unremarkable.
Hilar contours are similar compared to the prior exam.
Diffuse increased \textcolor{blue}{interstitial markings bilaterally suggest mild interstitial pulmonary edema.}
No pleural effusion or pneumothorax is identified.
No acute osseous abnormality seen. \\
\bottomrule
\end{tabularx}
\begin{tablenotes}
\footnotesize
\item \textcolor{blue}{Blue} indicates correct findings; \textcolor{red}{red} indicates incorrect/hallucinated statements; \textcolor{orange}{orange} indicates uncertain/partially wrong statements.
\end{tablenotes}
\end{threeparttable}
\end{table*}

\subsubsection{Summary of text-only results}
\emph{LLM2VEC}-style encoders exhibit a stronger grasp of report structure and
clinical semantics than BERT variants. Domain adaptation in
\emph{LLM2VEC4CXR} yields the best scores across most tasks. The 8B model achieves the highest absolute performance, while the fully fine-tuned 1B(\emph{LLM2VEC4CXR+}) offers a favorable accuracy--efficiency trade-off
(Appendix~\ref{app:compute_tradeoff}); we therefore adopt
\emph{LLM2VEC4CXR+} for multimodal experiments.

\subsection{Multimodal results}
Given that LLM-based text encoders outperform BERT-based models in text-only
tasks, we next examine whether \emph{LLM2VEC4CXR} can transfer its
domain-specific knowledge to the vision encoder within
\emph{LLM2CLIP4CXR}. \Cref{tab:mimic_openi} summarizes retrieval
performance and clinical evaluation metrics on the MIMIC (internal) and
Open-I (external) test sets.

\begin{table*}[h]
\centering
\caption{Multimodal image--text retrieval results on MIMIC and Open-I.}
\label{tab:mimic_openi}
\begin{threeparttable}
\resizebox{\textwidth}{!}{
\begin{tabular}{llcccccccccccc}
\toprule
& & \multicolumn{6}{c}{\textbf{MIMIC}} & \multicolumn{6}{c}{\textbf{Open-I}} \\
\cmidrule(lr){3-8} \cmidrule(lr){9-14}
\textbf{Model} & \textbf{Dataset} & \textbf{@1} & \textbf{@10} & \textbf{RadGraph} & \textbf{MF1} & \textbf{RaTE} & \textbf{GREEN} & \textbf{@1} & \textbf{@10} & \textbf{RadGraph} & \textbf{MF1} & \textbf{RaTE} & \textbf{GREEN} \\
\midrule
MAIRA2(7B)~\cite{maira2}     & M, P, U &  &  & 0.163 & 0.343 & 0.538 & 0.272  &  &  & 0.201  & 0.260 & 0.635 & 0.607\\
CheXagent(8B)~\cite{chen2024chexagent}   & M, C, P, B, Pu  &  &  & 0.154 & 0.248 & 0.510 & 0.256 &  &  & 0.154 & 0.186 & 0.604 & 0.539 \\
CXR-LLAVA~\cite{cxrllava}   & M, C, P, B &  &  & 0.092 & 0.116 & 0.487 & 0.143  &  &  & 0.187 & 0.052 & 0.559 & 0.357  \\
\hline
\hline
BC2C~\cite{bc2c}        & M & 0.076 & 0.336 & 0.133 & 0.374 & \textbf{0.512} & 0.235 & 0.017 & 0.081 & 0.148 & 0.179 & 0.554 & 0.483\\
CXR-CLIP~\cite{cxrclip}     & M & \textbf{0.175} & \textbf{0.553} & 0.138 & 0.342 & 0.494 & 0.232  & 0.029 & 0.101 & 0.133 & 0.171 & 0.537 & 0.421\\
GLoRIA~\cite{huang2021gloria}      & M & 0.027 & 0.149 & 0.106  & 0.293 & 0.477 & 0.163 & 0.011 & 0.079 & 0.129 & 0.128 & 0.505 & 0.381 \\
BioViL~\cite{biovil}      & M & 0.022 & 0.143 & 0.112  & 0.331 & 0.490 & 0.190  & 0.001 & 0.051 & 0.160 & 0.178 & 0.569 & 0.492 \\
BioViL-T~\cite{biovilt}    & M & 0.03 & 0.177 & 0.106  & 0.330 & 0.486 & 0.178 & 0.014 & 0.060 & 0.170 & 0.200 & 0.578 & 0.516 \\
MedCLIP~\cite{wang2022medclip}     & M & 0.001 & 0.011 & 0.041  & 0.287 & 0.403 & 0.078 & 0.002 & 0.012 & 0.031  & 0.112 & 0.309 & 0.053 \\
ConVIRT~\cite{convirt} & M & 0.114 & 0.437 & 0.130 & 0.353 & 0.506 & 0.232  & 0.019 & 0.085 & 0.148 & 0.165 & 0.558 & 0.482 \\
CLIP4CXR$_{\mathrm{base}}$    & M & 0.132 & 0.468 & 0.137 & 0.365 & 0.508 & 0.226 & 0.021 & 0.089 & 0.145 & 0.155 & 0.562 & 0.488\\
LLM2CLIP4CXR$_{\mathrm{base}}$  & M & 0.163 & 0.524 & \textbf{0.173} & \textbf{0.412} & 0.505 & \textbf{0.289} & \textbf{0.045} & \textbf{0.146} & \textbf{0.182} & \textbf{0.231} & \textbf{0.592} & \textbf{0.572} \\
\hline
BiomedCLIP~\cite{zhang2023biomedclip}  & M, Pu & 0.004 & 0.031 & 0.083  & 0.154 & 0.466 & 0.142  & 0.002 & 0.023 & 0.153 & 0.057 & 0.539 & 0.387\\
BC2C~\cite{bc2c}       & M,C & 0.068 & 0.311 & 0.105  & 0.382 & 0.515 & 0.237  & 0.019 & 0.083 & 0.141 & 0.191 & 0.562 & 0.486 \\
CXR-CLIP~\cite{cxrclip}     & M,C & 0.167 & 0.542 & 0.103  & 0.342 & 0.504 & 0.226 & 0.024 & 0.097 & 0.124 & 0.195 & 0.542 & 0.449 \\
CLIP4CXR$_{\mathrm{base}}$    & M,C & 0.107 & 0.429 & 0.099  & 0.354 & 0.508 & 0.242 & 0.014 & 0.077 & 0.146 & 0.172 & 0.575 & 0.495\\
LLM2CLIP4CXR$_{\mathrm{base}}$  & M,C & 0.181 & 0.559 & 0.178 & 0.422 & 0.538 & 0.299 & 0.048 & 0.153 & 0.187 & 0.228 & 0.601 & 0.600 \\
CLIP4CXR$_{\mathrm{base}}$    & M,C,P & 0.094 & 0.389 & 0.084  & 0.347 & 0.499 & 0.239 & 0.012 & 0.083 & 0.153 & 0.166 & 0.559 & 0.481\\
LLM2CLIP4CXR$_{\mathrm{base}}$  & M,C,P & 0.175 & 0.545 & 0.169 & 0.428 & 0.545 & 0.295 & 0.042 & 0.142 & 0.183 & 0.210 & 0.594 & 0.575 \\
\hline
CLIP4CXR$_{\mathrm{section}}$    & M,C,P & 0.091 & 0.385 & 0.081  & 0.350 & 0.497 & 0.232 & 0.010 & 0.085 & 0.159 & 0.168 & 0.545 & 0.479\\
LLM2CLIP4CXR$_{\mathrm{section}}$  & M,C,P & \textcolor{red}{\textbf{0.182}}  & 0.553 &  \textcolor{red}{\textbf{0.179}} & \textcolor{red}{\textbf{0.430}} &  \textcolor{red}{\textbf{0.547}} & 0.299 & \textcolor{red}{\textbf{0.052}} & 0.146 & 0.190 & \textcolor{red}{\textbf{0.231}} & 0.603 & 0.600 \\
\hline
CLIP4CXR$_{\mathrm{section}}$    & M,C,P, Pr & 0.072 & 0.352 & 0.058  & 0.322 & 0.483 & 0.227 & 0.006 & 0.052 & 0.144 & 0.152 & 0.523 & 0.452\\
LLM2CLIP4CXR$_{\mathrm{section}}$  & M,C,P, Pr & 0.179 & \textcolor{red}{\textbf{0.567}} & 0.178 & 0.417 & 0.546 & \textcolor{red}{\textbf{0.308}} & 0.049 & \textcolor{red}{\textbf{0.155}} & \textcolor{red}{\textbf{0.196}} & 0.230 & \textcolor{red}{\textbf{0.610}} & \textcolor{red}{\textbf{0.618}}\\
\bottomrule
\end{tabular}}
\begin{tablenotes}
\footnotesize
\item \textbf{Dataset abbreviations:} \textbf{M}=MIMIC, \textbf{C}=CheXpert-plus, \textbf{P}=PadChest, \textbf{U}=US-Mix, \textbf{B}=BimCV, \textbf{Pu}=other public datasets, \textbf{Pr}=private dataset.
\item \textbf{Highlighting:} \textbf{bold} indicates the best result among models trained only on MIMIC; \textcolor{red}{\textbf{red bold}} indicates the overall best result.
\item \textbf{Metric:} MF1 = CheXbert macro F1.
\end{tablenotes}
\end{threeparttable}
\end{table*}

\subsubsection{Comparison with CLIP baselines}
Among models trained solely on MIMIC, BC2C~\cite{bc2c} achieves strong clinical
metrics, while CXR-CLIP~\cite{cxrclip} obtains the highest Top-$k$ retrieval. On MIMIC,
\emph{LLM2CLIP4CXR} achieves comparable Top-$k$ retrieval but clearly exceeds
baselines in clinical efficacy measures, reflecting stronger alignment with
clinically meaningful content. On Open-I, it surpasses all baselines on both
Top-$k$ retrieval and clinical metrics. This substantial gain in external
validation underscores the advantage of LLM-based encoders, which provide
richer and more generalized text representations to guide the vision encoder.

\subsubsection{Effect of additional training data}\label{sec:moredata}
Several CLIP baselines show little gain—or even declines on MIMIC—after adding CheXpert-plus, likely due to formatting and style mismatches between sources. These results indicate that increasing data volume alone does not guarantee improved retrieval under report-style heterogeneity; robust text modeling is critical. In contrast, \emph{LLM2CLIP4CXR} maintains or improves performance on both MIMIC and Open-I, demonstrating that the LLM-based text encoder transfers cross-domain semantics more robustly. Importantly, our model can encounter reports written in diverse styles and still learn the underlying clinical facts efficiently, leading to stable or improved clinical alignment. Overall, clinically oriented metrics remain strong and are comparable to those reported by recent generative systems.

Adding PadChest has mixed effects: while some metrics improve, top-$k$
retrieval on MIMIC decreases, likely due to information density differences,
since PadChest reports typically contain only the impression. This discrepancy
is important because our private dataset (1.1M studies) also consists largely
of impression-only reports. To mitigate this, we introduce section placeholders
and instructions (\cref{sec:sectioned_processing}), producing
LLM2CLIP4CXR$_{\mathrm{section}}$. This variant better distinguishes between
\emph{Findings} and \emph{Impression}, mitigating performance drops and
improving external generalization. By contrast, the
CLIP4CXR$_{\mathrm{section}}$ baseline with BERT shows little benefit, further
underscoring the importance of stronger text encoders.

\subsubsection{Private dataset integration}
Finally, we augment training with a private corpus containing shorter,
abbreviated, impression-only reports. Since \emph{LLM2VEC4CXR} and
\emph{LLM2VEC} already demonstrate strong handling of abbreviations (see
\Cref{tab:llm2vec_results}), we expected them to adapt well. Indeed,
\emph{LLM2CLIP4CXR} maintains or slightly improves performance on MIMIC and
Open-I, while BERT-based models degrade more noticeably. This result highlights
the adaptability of LLM-based encoders to diverse and noisy clinical reporting
styles without compromising retrieval quality.

\subsubsection{\rev{Summary of multimodal retrieval}}
\rev{Overall, \emph{LLM2CLIP4CXR} scales robustly across diverse reporting
styles: as stylistically varied sources are progressively added, its clinical
retrieval performance is maintained or even improved, particularly on external
Open-I evaluation (\Cref{tab:mimic_openi}). BERT-based CLIP baselines instead
degrade as report styles diverge, showing that a robust LLM-based text encoder,
rather than data volume alone, is what enables clinically faithful and
generalizable retrieval.}

\subsection{Qualitative evaluation of retrieved reports}
As noted in \cref{sec:experiments}, automated metrics may overlook clinically important
aspects of report quality. To complement them, we performed a qualitative evaluation
on 200 Open-I cases using three medical-student raters and four LLM judges. We included
the generative model \emph{MAIRA2} as a reference system and retrained four CLIP-based
retrieval models (\emph{CLIP4CXR}, \emph{CXR-CLIP}, BC2C~\cite{bc2c}, and
\emph{LLM2CLIP4CXR$_{\mathrm{MC}}$}) under identical MIMIC+CheXpert conditions
for fair comparison. We also evaluated our large-scale model,
\emph{LLM2CLIP4CXR$_{\mathrm{all}}$}, trained on 1.6M pairs.

\Cref{tab:qual_summary} shows that LLM2CLIP4CXR$_{\mathrm{all}}$
achieves the best average rank among both student raters and LLM judges.
Among models trained under the same MIMIC+CheXpert setting,
\emph{LLM2CLIP4CXR$_{\mathrm{MC}}$} achieves the best qualitative performance
among CLIP-based retrieval models and ranks favorably relative to MAIRA2.
These findings suggest that LLM-based encoders improve the clinical relevance
of retrieved reports while preserving the controllability of retrieval-based output. \rev{The thoracic radiologist's rankings on the 72-case subset are summarized
in \Cref{tab:rad_subset}, alongside the corresponding student and LLM-judge
rankings on the same cases.}

\begin{table}[h]
\centering
\caption{Qualitative evaluation on 200 Open-I cases. Student values
are averaged across three medical-student raters, and LLM values are averaged across four LLM judges.}
\label{tab:qual_summary}
\begin{threeparttable}
\resizebox{\linewidth}{!}{
\begin{tabular}{lrrr}
\toprule
\textbf{Model} & \textbf{Student Mean Rank} $\downarrow$ & \textbf{LLM Mean Rank} $\downarrow$ & \textbf{Overall Mean Rank} $\downarrow$\\ 
\midrule
\emph{LLM2CLIP4CXR}$_{\mathrm{all}}$\tnote{a} & \textbf{1.85} & \textbf{2.46} & \textbf{2.19}\\
\emph{LLM2CLIP4CXR}$_{\mathrm{MC}}$  & 2.46 & 2.75 & 2.62\\
MAIRA2~\cite{maira2}          & 2.50 & 3.55 & 3.08\\
CLIP4CXR                      & 3.54 & 4.00 & 3.80\\
CXR-CLIP~\cite{cxrclip}       & 3.71 & 4.22 & 4.03\\
BC2C~\cite{bc2c}              & 3.47 & 3.89 & 3.71\\
\bottomrule
\end{tabular}}
\begin{tablenotes}
\footnotesize
\item[a] \emph{LLM2CLIP4CXR}$_{\mathrm{all}}$ denotes the model trained on the full training corpus, \\ including MIMIC-CXR, CheXpert-plus, PadChest, and SNUH reports.
\item Lower rank indicates better qualitative performance. Ties were permitted.
\end{tablenotes}
\end{threeparttable}
\end{table}

\begin{table*}[h]
\centering
\caption{\rev{Thoracic-radiologist-reference ranking on the 72-case subset. Ranks were computed within the three representative systems only; lower mean rank indicates better clinical match to the ground-truth report.}}
\label{tab:rad_subset}
\resizebox{\textwidth}{!}{
\begin{tabular}{lcccccc}
\toprule
\textbf{Model} 
& \textbf{Expert mean rank} $\downarrow$
& \textbf{Expert top-rank votes} $\uparrow$
& \textbf{Student mean rank} $\downarrow$
& \textbf{Student top-rank votes} $\uparrow$
& \textbf{LLM mean rank} $\downarrow$
& \textbf{LLM top-rank votes} $\uparrow$ \\
\midrule
LLM2CLIP4CXR$_{\mathrm{all}}$ 
& \textbf{1.29} & \textbf{55/72 (76\%)} 
& \textbf{1.14} & \textbf{193/216 (89\%)} 
& \textbf{1.43} & \textbf{191/288 (66\%)} \\
MAIRA2~\cite{maira2}
& 2.21 & 11/72 (15\%) 
& 2.07 & 62/216 (29\%) 
& 2.39 & 48/288 (17\%) \\
BC2C~\cite{bc2c}
& 2.50 & 6/72 (8\%) 
& 2.35 & 35/216 (16\%) 
& 2.17 & 53/288 (18\%) \\
\bottomrule
\end{tabular}
}
\end{table*}

\subsubsection{\rev{Summary of qualitative evaluation}}
\rev{Across all rater types \emph{LLM2CLIP4CXR$_{\mathrm{all}}$} was consistently judged to
capture clinical semantics more faithfully than the baselines
(\Cref{tab:qual_summary,tab:rad_subset}), with the expert ranking it first in
76\% of cases. This agreement, holding even under specialist radiologist review,
indicates that the LLM-based encoder encodes richer and more clinically accurate
report semantics; notably, the retrieved reports were also judged comparable to
or better than those from a report-generation baseline.}

\section{Discussion}

This study demonstrates that incorporating LLM-based encoders into medical vision--language frameworks substantially improves clinical text understanding and cross-dataset generalization. By developing domain-adapted LLM encoders that handle the linguistic diversity of radiology reports, including abbreviated, templated, and impression-only styles, our approach enables more consistent multimodal learning across heterogeneous datasets.

\rev{Together, the text-only, multimodal, and dataset-composition experiments address the central question posed in the Introduction: whether stronger report representations can improve medical VLP under heterogeneous reporting styles. The results show that domain-adapted LLM encoders better capture CXR report semantics than BERT-based encoders and that this advantage transfers to CLIP-style retrieval, particularly when training data include impression-only or abbreviation-heavy reports.}

The proposed \emph{LLM2VEC4CXR} and \emph{LLM2CLIP4CXR} frameworks further show that clinically robust medical VLP depends not only on dataset scale, but also on how report semantics are encoded. Combining LLM-based embeddings with masked token prediction and contrastive learning improves robustness to stylistic variation and supports clinically faithful retrieval, as reflected by gains on the error-discrimination and clinical-similarity evaluations. Importantly, our results indicate that large-scale integration of private, unstructured hospital reports is feasible when the text encoder is designed to handle report-style heterogeneity.

This study has several limitations. First, although we evaluated external generalization on Open-I and included heterogeneous public and private training data, the evaluation may not fully capture reporting variation across institutions, scanners, and clinical workflows. Second, \rev{expert validation was performed by a single thoracic radiologist, so inter-radiologist variability and consensus reliability could not be assessed.} Third, our experiments focus on CXRs, and extending the framework to other imaging modalities will be necessary to assess broader generalizability. Finally, image-to-text retrieval serves as a proxy for multimodal understanding and does not directly evaluate downstream clinical decision-making or prospective workflow impact.

\rev{Clinically, LLM2CLIP4CXR could serve as a retrieval layer within PACS/RIS workflows by surfacing prior or similar CXR cases with semantically matched reports, supporting report drafting, discrepancy review, and image–report consistency checking. Because the system retrieves existing image–report pairs rather than generating unconstrained diagnostic text, its outputs remain traceable and can be reviewed by radiologists as decision support; however, prospective workflow integration, latency testing, and safety monitoring are required before clinical deployment.}

Future work will address these limitations through multi-institutional, style-stratified, and temporal validation with multiple radiologists. \rev{The framework will be extended beyond CXRs to other imaging modalities and multi-view or longitudinal settings; \rev{in particular, the domain-adapted encoder (\emph{LLM2VEC4CXR}) could be transferred to clinically adjacent thoracic report domains such as chest CT, with appropriate domain-specific adaptation, to probe its capability beyond CXR.}} To move beyond retrieval as a proxy task, we will evaluate downstream clinical applications enabled by retrieval-based representations, including report--image mismatch detection, similar-case retrieval, and clinician-in-the-loop decision support.

\section{Conclusion}
\label{sec:conclusion}

We introduced \emph{LLM2VEC4CXR}, a domain-adapted LLM encoder for chest X-ray reports, and its multimodal extension \emph{LLM2CLIP4CXR} for image--text retrieval. Across diverse benchmarks, these models consistently outperformed BERT-based and prior medical CLIP variants, capturing clinical semantics more effectively and maintaining robustness across heterogeneous, abbreviation-rich, and impression-focused reporting styles. Trained on 1.6M CXRs from public and private sources, \rev{and supported by expert thoracic-radiologist validation,} our models demonstrate that LLM-based encoders enable scalable, clinically aware, and generalizable multimodal learning, offering a foundation for more robust medical vision--language models in future research.

\section*{Ethics approval}
Use of public datasets (MIMIC-CXR, CheXpert-plus, PadChest, Open-I) complied with their data-use agreements and institutional policies; all data are fully de-identified. For the private hospital dataset, this retrospective study was approved by the Institutional Review Board of Seoul National University Hospital (IRB No.: 2405-070-1536), which granted a waiver of informed consent due to the use of de-identified data. No prospective data collection or intervention was performed.

\section*{Code availability}
We release model checkpoints and code for research use. The \textit{LLM2VEC4CXR} model is available at \url{https://huggingface.co/lukeingawesome/llm2vec4cxr} and \url{https://github.com/lukeingawesome/llm2vec4cxr}, and training/inference code for \textit{LLM2CLIP4CXR} is available at \url{https://github.com/lukeingawesome/llm2clip4cxr}. Please see the repositories for licenses and instructions.

\newpage
\appendices

\section{Datasets}

We summarize dataset usage and provide additional details of the preprocessing steps.

\subsection{Use of Indication fields for abbreviation learning}
To improve robustness to shorthand, we incorporate \emph{Indication}
fields from MIMIC, which frequently contain abbreviations and condensed
expressions. These are paired with expanded versions created through our
variation pipeline. This pairing enables the encoder to align abbreviated inputs with their
clinically complete forms, strengthening its ability to handle style variation.

\subsection{Prompt design for report variation generation}
\label{sec:prompts}

We used large language models to generate clinically faithful report variants from MIMIC-CXR and CheXpert data. For MIMIC-CXR, preprocessing was conducted with Gemini~2.0-Flash via Vertex~AI, in accordance with the responsible-use guidelines published on PhysioNet\footnote{\url{https://physionet.org/news/post/gpt-responsible-use}}. The preprocessing was completed prior to September~24, 2025, before the release of updated guidelines. For CheXpert, we applied DeepSeek-R1-Distill-Qwen-14B locally to ensure controlled data handling and compliance with institutional data governance.

\paragraph{Splitting and omission}
Sentence splitting and prior-omission were implemented as described by
~\cite{bc2c}.

\paragraph{Anatomical partitioning}
We used the prompts in RadExtract~\cite{radextract2024} that instructed the model to
decompose each report into separate anatomical regions (e.g., pleura, lung zones,
mediastinum). Recombination into structured variants was rule-based for each anatomy part.

\paragraph{Others}
For error generation, creation of clinically similar reports, and evaluation we report the prompts in the github repository \url{https://github.com/lukeingawesome/llm2vec4cxr}

\subsection{Prompt templates for LLM2CLIP4CXR training}
\label{app:prompt_templates}

\rev{
\begin{table}[t]
\centering
\caption{Instruction prompt templates used for LLM2CLIP4CXR training.}
\label{tab:prompt_templates}
\footnotesize
\begin{tabular}{p{0.28\linewidth}p{0.64\linewidth}}
\toprule
\textbf{Task type} & \textbf{Prompt template} \\
\midrule
Semantic similarity &
\texttt{Retrieve semantically similar sentences.} \\
Summarization &
\texttt{Summarize the CXR report.} \\
Finding-level classification &
\texttt{Determine the change or status of the \{finding\}.} \\
\bottomrule
\end{tabular}
\end{table}
}
\rev{
Table~\ref{tab:prompt_templates} summarizes the fixed instruction prompts used during LLM2CLIP4CXR training. The prompts were designed to cover complementary report-understanding objectives, including semantic similarity, summarization, and finding-level classification. The same templates were used throughout training and were not tuned on individual evaluation cases.
}

\section{Additional Experiments}

\subsection{Computational trade-off}
\label{app:compute_tradeoff}

\begin{table}[!h]
\centering
\caption{\rev{Computational cost of LLM2VEC4CXR variants. Peak GPU memory and step time are reported under the same four-A6000 training setup.}}
\label{tab:compute_tradeoff}
\resizebox{\linewidth}{!}{
\begin{tabular}{lccc}
\toprule
\textbf{Configuration} 
& \textbf{Trainable parameters} 
& \textbf{Peak GPU memory} 
& \textbf{Step time} \\
\midrule
1B + LoRA $(r=16)$ 
& 16M (1.3\%) 
& $\sim$10 GiB 
& $\sim$6 s \\
1B full FT 
& 1{,}236M (100\%) 
& $\sim$14 GiB 
& $\sim$7 s \\
8B + LoRA $(r=16)$ 
& 59M (0.7\%) 
& $\sim$31 GiB 
& $\sim$44 s \\
\bottomrule
\end{tabular}
}
\end{table}

\rev{\Cref{tab:compute_tradeoff} show that LoRA does not eliminate the computational burden of a large backbone. While 8B LoRA provides the strongest text-encoder capacity, the 1B fully fine-tuned model offers a more favorable accuracy–efficiency balance, requiring substantially lower memory and step time while retaining competitive downstream performance. We therefore use the 1B setting as the more practical configuration for scalable multimodal training.}

\subsection{Additional retrieval examples for private dataset}
\Cref{tab:retrieval_example_private} shows retrieval examples comparing 
\emph{LLM2VEC4CXR} with the original \emph{LLM2VEC} and BERT-based encoders, 
using private reports with frequent abbreviations as queries and MIMIC reports 
as candidates. BERT-based models and the general \emph{LLM2VEC} often fail to 
interpret these abbreviations and to retrieve the correct clinical findings. 
In contrast, \emph{LLM2VEC4CXR} accurately expands the abbreviations and 
retrieves clinically similar reports, demonstrating stronger robustness to 
abbreviated and noisy reporting styles.

\begin{table*}[h]
\centering
\footnotesize
\caption{Top-1 retrieved report for Task~5. A report from private dataset is used as the query,
and the model retrieves the most similar report from the MIMIC pool. 
\textcolor{blue}{Blue} = correct findings; \textcolor{red}{red} = incorrect/hallucinated; 
\textcolor{orange}{orange} = uncertain/partially wrong statements.}
\label{tab:retrieval_example_private}
\setlength{\tabcolsep}{6pt}
\renewcommand{\arraystretch}{1.15}
\begin{tabularx}{\textwidth}{@{} l X @{}}
\toprule
\textbf{Model} & \textbf{Text} \\
\midrule
Ground Truth (Query) & \textcolor{blue}{A tiny calcified granuloma, RULF.}  No active lung disease. \\
\hline
\hline
Base BERT & In comparison with the study of \_, the \textcolor{red}{heart remains at the upper limits of normal in size and there is some tortuosity of the aorta}. No vascular congestion. No evidence of acute focal pneumonia at this time. \\
\hline
ClinicalBERT & In comparison with the study of \_, there is no evidence of acute cardiopulmonary disease at this time. No vascular congestion, pleural effusion, or acute pneumonia. \textcolor{red}{No evidence of old granulomatous disease.} \\
\hline
LLM2VEC(1B) & \textcolor{red}{Mild cardiomegaly. Lung volumes are low.} There is no focal consolidation. No pneumothorax. \\
\hline
LLM2VEC4CXR+ & \textcolor{blue}{There is a small calcified granuloma in the right upper lung field.} This is of no clinical significance. There is no lung consolidation, pleural effusion, or pneumothorax. Cardiomediastinal silhouette and hilar structures are normal. \\
\bottomrule
\end{tabularx}
\end{table*}

\begin{table*}[h]
\centering
\caption{
Ablation of dataset composition for training \emph{LLM2VEC4CXR+}.}
\label{tab:ablation_loso}
\resizebox{\textwidth}{!}{
\begin{tabular}{lccccccccccc}
\toprule
{\textbf{Excluded subset}} &
\multicolumn{2}{c}{\textbf{Task 1} ($\uparrow$)} &
\multicolumn{2}{c}{\textbf{Task 2} ($\uparrow$)} &
{\textbf{Task 3} ($\uparrow$)} &
\multicolumn{2}{c}{\textbf{Task 4} ($\uparrow$)} &
\multicolumn{3}{c}{\textbf{Task 5} ($\uparrow$)} \\
\cmidrule(lr){2-3}\cmidrule(lr){4-5}\cmidrule(lr){7-8}\cmidrule(lr){9-11}
 & @1 & @5 & @1 & @5 & Acc & @1 & @3 & RadGraph & MF1 & GREEN \\
\midrule
\textbf{Full (no exclusion)} & 0.918 & 0.984 & 0.203 & 0.290 & 0.826 & 0.607 & 0.854 & 0.373 & 0.715 & 0.686 \\
\midrule
-- Summarization reports          & 0.912 & 0.978 & 0.115 & 0.182 & 0.812 & 0.611 & 0.854 & 0.353 & 0.698 & 0.668 \\
-- Prior-omitted reports          & 0.867 & 0.899 & 0.182 & 0.271 & 0.812 & 0.594 & 0.820 & 0.361 & 0.664 & 0.658 \\
-- Anatomical-partitioned reports & 0.901 & 0.912 & 0.196 & 0.278 & 0.790 & 0.607 & 0.795 & 0.386 & 0.712 & 0.662 \\
-- Clinically similar reports     & 0.910 & 0.978 & 0.156 & 0.206 & 0.728 & 0.417 & 0.783 & 0.342 & 0.652 & 0.646 \\
\bottomrule
\end{tabular}
}
\end{table*}

\begin{table*}[h]
\centering
\caption{Ablation of pooling strategy for training \emph{LLM2VEC4CXR+}.}
\label{tab:ablation_pooling}
\resizebox{\textwidth}{!}{
\begin{tabular}{lccccccccccc}
\toprule
{\textbf{Pooling method}} &
\multicolumn{2}{c}{\textbf{Task 1} ($\uparrow$)} &
\multicolumn{2}{c}{\textbf{Task 2} ($\uparrow$)} &
{\textbf{Task 3} ($\uparrow$)} &
\multicolumn{2}{c}{\textbf{Task 4} ($\uparrow$)} &
\multicolumn{3}{c}{\textbf{Task 5} ($\uparrow$)} \\
\cmidrule(lr){2-3}\cmidrule(lr){4-5}\cmidrule(lr){7-8}\cmidrule(lr){9-11}
 & @1 & @5 & @1 & @5 & Acc & @1 & @3 & RadGraph & MF1 & GREEN \\
\midrule
\textit{Latent attention} & 0.918 & 0.984 & 0.203 & 0.290 & 0.826 & 0.607 & 0.854 & 0.373 & 0.715 & 0.686 \\
\textit{Mean pooling}     & 0.901 & 0.960 & 0.188 & 0.271 & 0.812 & 0.556 & 0.762 & 0.381 & 0.704 & 0.670 \\
\bottomrule
\end{tabular}
}
\end{table*}

\subsection{Ablation}
\paragraph{Ablation for dataset composition}
We conducted ablation experiments to analyze the effect of dataset composition during the training of \emph{LLM2VEC4CXR}. As shown in Table~\ref{tab:ablation_loso}, each component contributed meaningfully to overall performance, indicating that exposure to diverse linguistic patterns enhances text encoder robustness. Among them, the inclusion of clinically similar reports produced the largest gains, reflecting their role in learning medical abbreviations and integrating stylistically varied report structures. In addition, we evaluated different text-pooling strategies (Table~\ref{tab:ablation_pooling}) and found that latent attention pooling outperformed simple mean pooling, suggesting that selectively weighting token importance yields more informative clinical representations.

% \section*{Declaration of competing interest}
% The authors declare no competing interests.

\section*{Acknowledgment}
This research was supported by a grant from the Korea Health Technology R\&D Project through the Korea Health Industry Development Institute (KHIDI), funded by the Ministry of Health \& Welfare, Republic of Korea (Grant No. RS-2025-02213531; NTIS No. 2460004432). This study was also supported by the National Research Foundation of Korea (NRF) grant funded by the Ministry of Science and ICT (MSIT) (Grant No. RS-2024-00354666).

Generative AI was used in limited, disclosed ways: large language models generated synthetic training-report variants, served as judges in the qualitative evaluation (Methods and Experiments), and assisted with grammar refinement. All scientific content, analyses, and conclusions were produced and verified by the authors, who take full responsibility for the work.

\section*{References}
\bibliographystyle{IEEEtran}
\bibliography{jbhi}

\end{document}